\documentclass{article} 
\usepackage[preprint]{colm2026_conference}

\usepackage{microtype}
\usepackage{hyperref}
\usepackage{url}
\usepackage{booktabs}
\usepackage{graphicx}

\usepackage{placeins}
\usepackage{hyperref}
\usepackage{url}
\usepackage{graphicx}

\usepackage[utf8]{inputenc}
\usepackage{xcolor} 
\usepackage{amsmath}
\usepackage{amsfonts}
\usepackage{dcolumn} 
\usepackage{tabu}
\usepackage{array}
\usepackage{xspace}
\usepackage{makecell}
\usepackage{amsthm}

\usepackage[table]{xcolor}

\usepackage{enumitem}
\usepackage{algorithm}
\usepackage{algorithm}
\usepackage{colortbl}
\usepackage{xspace}
\usepackage{booktabs, multirow} 
\usepackage{soul}
\usepackage{changepage,threeparttable} 

\usepackage{hyperref}
\usepackage{url}
\usepackage{array}
\usepackage{booktabs}       
\usepackage{amsfonts}       
\usepackage{nicefrac}       
\usepackage{microtype}      
\usepackage{xcolor}         
\definecolor{LightBlue}{HTML}{E8F4FD}
\usepackage{subcaption}
\definecolor{darkblue}{rgb}{0, 0, 0.5}
\hypersetup{colorlinks=true, citecolor=darkblue, linkcolor=darkblue, urlcolor=darkblue}

\usepackage{wrapfig}
\usepackage{overpic}
\usepackage{placeins}
\usepackage{algpseudocode}
\usepackage{multirow}
\usepackage{graphicx}
\usepackage{url}
\usepackage{soul}  
\usepackage{marginnote}  

\newcommand{\vd}{\mathbf{d}}
\newcommand{\vh}{\mathbf{h}}

\newcommand{\textdnewline}{\texttt{\textbackslash n\textbackslash n}}
\newcommand{\textnewline}{\texttt{\textbackslash n}}

\definecolor{verylightblue}{RGB}{235,242,255}
\definecolor{mediumblue}{RGB}{0, 100, 200}
\definecolor{verylightred}{RGB}{255,239,239}

\usepackage[breakable,skins]{tcolorbox}
\usepackage{listings}
\setulcolor{red} 
\makeatletter
\def\SOUL@ulthickness{1.5pt}  
\def\SOUL@uldepth{2pt}        
\makeatother

\definecolor{questionblue}{RGB}{0, 100, 200}
\definecolor{goldyellow}{RGB}{184, 134, 11}
\definecolor{llmgreen}{RGB}{0, 150, 0}
\definecolor{judgmentpurple}{RGB}{128, 0, 128}
\definecolor{highlightyellow}{RGB}{255, 255, 150}
\definecolor{errorred}{RGB}{255, 200, 200}
\definecolor{darkgreen}{RGB}{34, 139, 34}  
\usepackage{tikz}

\newcommand{\strictv}{\vd^l_{\text{strict}}}

\newcommand{\ourmethod}{\texttt{VerifySteer}\xspace}
\newcommand{\ourmethoduni}{\texttt{VerifySteer-Uni}\xspace}
\newcommand{\ourmethodbi}{\texttt{VerifySteer-Bi}\xspace}
\newcommand{\strictvwol}{\vd_{\text{strict}}}
\newcommand{\lenientvwol}{\vd_{\text{lenient}}}

\usepackage{amsmath,amsfonts,bm,amssymb}

\def\eqref#1{equation~\ref{#1}}

\def\1{\bm{1}}

\def\vd{{\bm{d}}}

\def\vh{{\bm{h}}}

\DeclareMathAlphabet{\mathsfit}{\encodingdefault}{\sfdefault}{m}{sl}
\SetMathAlphabet{\mathsfit}{bold}{\encodingdefault}{\sfdefault}{bx}{n}

\usepackage{lineno}
\usepackage{cleveref}
\crefname{appendix}{appendix}{appendices}
\Crefname{appendix}{Appendix}{Appendices}

\title{The Hidden Signal of Verifier Strictness: Controlling and Improving Step-Wise Verification via Selective Latent Steering}

\newcommand*\samethanks[1][\value{footnote}]{\footnotemark[#1]}
\author{%
  Yefan Zhou\textsuperscript{1}, Yilun Zhou\textsuperscript{2}, Austin Xu\textsuperscript{3}\thanks{Work done at Salesforce AI Research}, 
  Soroush Vosoughi\textsuperscript{1}\thanks{Equal advising},
  Shafiq Joty\textsuperscript{3}\samethanks[2],
  Jiang Gui\textsuperscript{1}\samethanks[2]
  \\
  {\textsuperscript{1}Dartmouth College, \textsuperscript{2}Datadog AI Research, \textsuperscript{3}Salesforce AI Research} \\
}

\begin{document}

\ifcolmsubmission
\linenumbers
\fi

\maketitle

\begin{abstract}
Generative verifiers have emerged as a promising paradigm for step-wise verification, but their verification behavior is often poorly calibrated: they may be under-critical and miss erroneous steps, or over-critical and reject correct reasoning.  
We refer to this tendency to be overly lenient or overly critical as verifier strictness.
In this work, we study whether verifier strictness can be controlled through hidden-state intervention. 
We uncover a verification-specific hidden-state signal: in step-wise verification, a verifier’s tendency to accept or reject a solution step is encoded near the boundary of the corresponding verification paragraph. 
Exploiting this signal, we show that hidden-state steering can directly modulate verifier strictness without fine-tuning. 
However, uniform steering induces a trade-off between error detection and correctness certification.
To address this, we propose \ourmethod, which exploits latent correctness signals for sample-level routing and selectively intervenes on paragraph boundaries.
Experiments on ProcessBench and Hard2Verify show that \ourmethod outperforms prompt optimization and activation steering baselines, and is competitive with self-consistency while requiring $4$--$7\times$ less inference compute. 
\ourmethod{} is also complementary to verification fine-tuning, providing further gains on top of fine-tuned verifiers. The code is available at \url{https://github.com/YefanZhou/VerifySteer}.
\looseness-1
\end{abstract}

\section{Introduction}

Large Language Models (LLMs) have made remarkable progress in mathematical reasoning. 
Yet they still make notable errors, including miscalculations and logical mistakes that lead to incorrect conclusions~\citep{yin-etal-2025-error}.  
Crucially, these errors do not appear only in the final answer. 
Recent work shows that LLMs can also make step-level mistakes in intermediate reasoning, sometimes even while arriving at the correct final answer~\citep{zheng2025processbench, setlur2025rewarding}. 
This challenge motivated step-level verification and Process Reward Models (PRMs; \citealp{wang2024math, lightman2023let}) with the aim of detecting errors at the level of reasoning steps.
More recently, attention has shifted to \emph{generative verifiers}~\citep{zhang2025generative, mahan2024generative, liu2025inference}. 
These models treat verification as a next-token prediction problem. 
Given a problem and a candidate solution, a generative verifier typically generates a verification chain-of-thought (CoT) and then outputs a verdict.
This paradigm can be directly applied to general instruction-tuned LLMs through prompting, and provides an interpretable CoT for explaining the step-level mistake. 
\looseness-1

\begin{figure*}[!th]
\centering
    \includegraphics[width=0.95\linewidth]{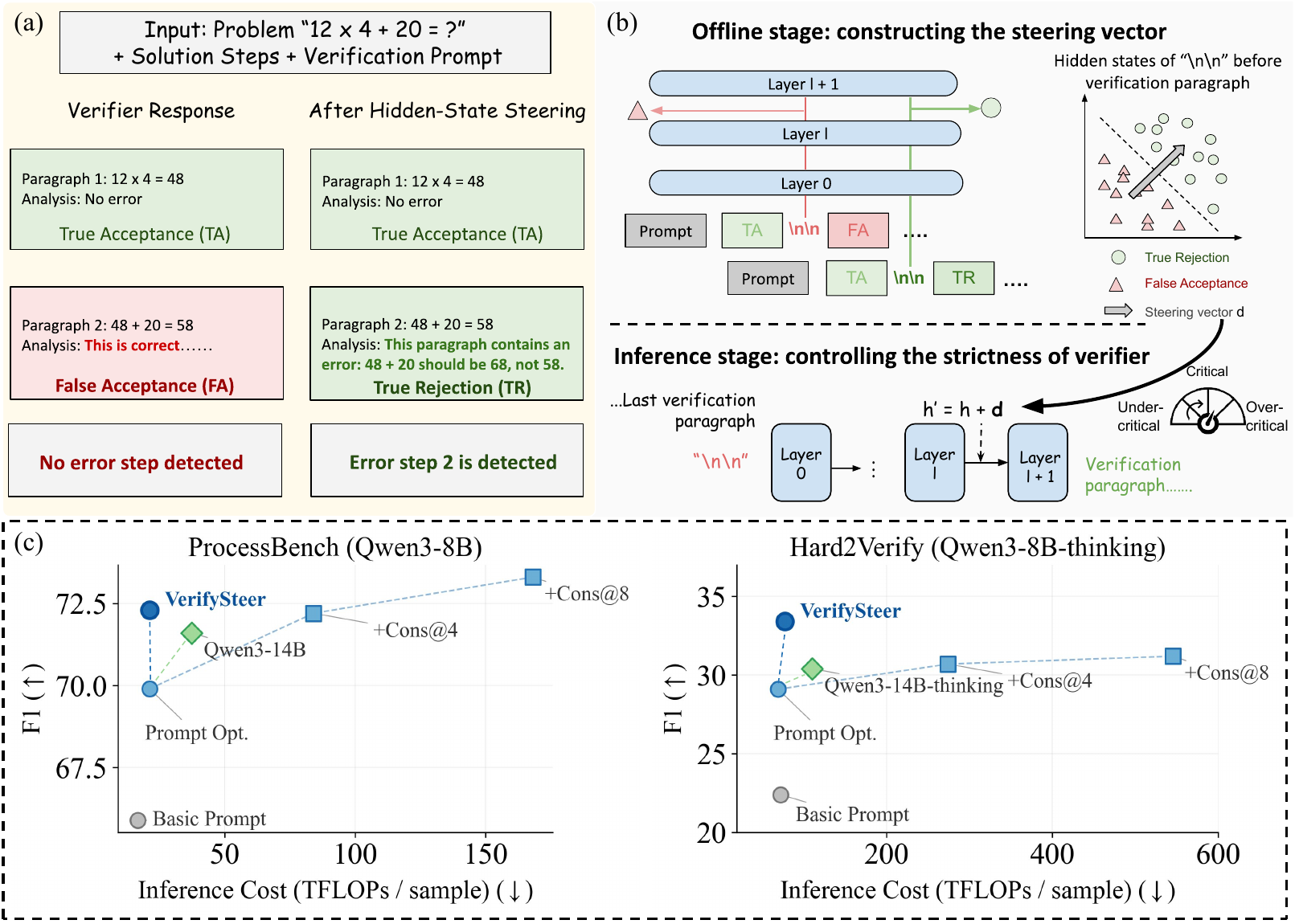}\vspace{-2mm}
    \caption{\textbf{Overview of our findings on controlling step-wise verification through hidden-state steering.}
    (a) Effect of steering.
    The baseline verifier falsely accepts an erroneous step and misses the true error location.
    After steering, the verifier correctly rejects the erroneous step and identifies the exact error location.
    (b) Steering pipeline.
    In the offline stage, we collect hidden states of the paragraph-boundary delimiter token (\texttt{\textbackslash n\textbackslash n}) at transformer layer $l$ on a validation set. 
    TA, FA, and TR denote true acceptance, false acceptance, and true rejection verification paragraphs, respectively.
    We construct a steering vector $\mathbf{d}$ from the mean difference between true rejection (TR) and false acceptance (FA) states.
    At inference time, given an unseen sample to verify, we add the steering vector ($\mathbf{d}$) to the delimiter-token hidden states at layer $l$ to shift verifier strictness.
    (c) Performance-cost tradeoffs on ProcessBench and Hard2Verify for our method \ourmethod and baselines.
    \texttt{Cons@4} and \texttt{Cons@8} denote self-consistency decoding with 4 and 8 sampled verification traces respectively.
    \looseness-1
    }\vspace{-3mm}
    \label{fig:obs-demo}
\end{figure*}

Recent benchmark and evaluation studies have revealed important limitations of generative verification. 
A recurring pattern is a \textit{positivity bias}, where the verifier tends to incorrectly accept flawed solutions.
This bias arises in several settings, including self-verification and intra-family verification~\citep{lu2025does, singh2026v_1}, as well as when a weaker verifier evaluates solutions produced by a stronger generator~\citep{pandit2025hard2verify, zhou2026variation}.
Although fine-tuning specialized verifiers~\citep{xiong2025stepwiser, zha2025rl, xu2026foundational} could potentially mitigate this bias, there is strong motivation to investigate training-free methods that can be directly applied to general instruction-tuned models.
A key reason is the rapid iteration of both open-source and closed-source frontier models, whose reasoning ability on solving complex problems keeps improving over the span of just a few months. 
Since verification ability is shown to be positively correlated with problem-solving ability~\citep{tan2024judgebench, chen2025llm, krumdick2025no}, specialized verifiers trained on older model generations can quickly become outdated and may underperform newer instruction-tuned models.
Indeed, \citet{pandit2025hard2verify} show that recent instruction-tuned models such as Qwen3~\citep{yang2025qwen3} already outperform specialized fine-tuned step-wise evaluators (e.g., Qwen2.5-Math-PRM-72B~\citep{zhang2025lessons}) that are substantially larger in size.
Moreover, fine-tuning for step-wise verification is expensive in both training and data curation: the Qwen-Math PRMs require 860K training examples, and the step-level annotation pipeline~\citep{duan2025efficient} consumes an estimated 17B tokens from an LLM judge.
Existing evaluator pipelines also require solution traces from current generator models, yet \citet{singh2026on} show that evaluators trained on earlier generations generalize poorly to responses from new models.
These observations motivate a training-free methodology that can adapt efficiently to evolving general instruction-tuned models.\looseness-1

Drawing inspiration from recent work in representation engineering~\citep{zou2023representation, wu2025axbench, maar2026whats, dong2025emergent, kumaran2026llms}, which suggests that LLM behaviors are encoded in hidden states and can be controlled through targeted interventions, we study whether hidden-state intervention can be used to control step-wise verification in a training-free manner.
Our key finding is that the verifier's tendency to accept or reject a solution step is encoded in the paragraph-boundary delimiter token preceding the corresponding verification paragraph.
By intervening on this hidden state at a selected transformer layer, we can directly modulate verifier strictness.
An overview of this finding is illustrated in \Cref{fig:obs-demo}.
Concretely, this basic intervention applies a contrastive steering vector uniformly to every sample and every generated delimiter token, nudging the verifier toward stricter or more lenient judgment depending on the steering direction.
As shown in our experiments, this uniform steering improves either error detection or correctness certification, while preserving semantic coherence and turning superficial verification CoT into detailed explanations of mistakes.

However, uniform steering exposes a trade-off: increasing strictness tends to improve error detection at the cost of correctness certification, while decreasing strictness has the opposite effect.
To address this, we propose \ourmethod, which introduces both sample-level and delimiter-level adaptivity to control verifier strictness.
At the sample level, \ourmethod uses a lightweight correctness probe to predict whether the candidate solution is likely fully correct, and routes each sample to the appropriate steering direction based on this score.
This design is motivated by recent work~\citep{liang2026hidden, ye2025physics, zhang2025reasoning, li2026rethinking} showing that hidden states in the middle layers encode rich signals about answer correctness and can support verification.
At the delimiter level, \ourmethod intervenes only on paragraph-boundary tokens whose hidden states are misaligned with the selected direction.
Experiments on ProcessBench and Hard2Verify show that \ourmethod achieves the highest F1 at comparable inference cost to single-pass baselines and a larger 14B verifier (\Cref{fig:obs-demo}(c)), and is competitive with self-consistency while requiring $4$--$7\times$ fewer FLOPs.
\looseness-1

Our key contributions are as follows: 
(i) We identify a verification-specific hidden-state signal: the verifier’s tendency to accept or reject a solution step is encoded in the hidden state of the paragraph-boundary delimiter token, and steering this signal can modulate verifier strictness.
(ii) Building on this finding, we propose \ourmethod{}, which introduces sample-level and delimiter-level adaptivity to control verifier strictness, with ablations confirming the importance of each component.
(iii) Experiments on ProcessBench and Hard2Verify show that \ourmethod{} outperforms prompt optimization, activation steering~\citep{rimsky-etal-2024-steering}, and self-consistency~\citep{wang2023selfconsistency} baselines.
For example, on ProcessBench with Qwen3-1.7B~\citep{yang2025qwen3}, \ourmethod{} surpasses self-consistency by 3.7 F1 while using $4\times$ less inference compute. On Hard2Verify with Qwen3-8B-thinking, \ourmethod{} surpasses self-consistency by 2.2 F1 while using $7\times$ less compute.
Our evaluation spans a variety of models, with scales ranging from 1.7B to 20B parameters, dense and mixture-of-experts (MoE) architectures, and both non-thinking and thinking modes.
\ourmethod{} is also orthogonal to and compatible with verification fine-tuning, providing further gains when applied on top of FARE-20B~\citep{xu2026foundational}, a verifier fine-tuned from gpt-oss-20B~\citep{openai2025gptoss120bgptoss20bmodel}.\looseness-1

\vspace{-3mm}
\section{Related Work}\label{sec:related_work}
\vspace{-3mm}

\textbf{LLM Verification.} Generative verifiers have emerged as an evaluation paradigm comparable to scalar reward models~\citep{lightman2023let, zhang2025lessons}. 
They can be broadly categorized into step-level verification~\citep{zheng2025processbench} and sequence-level verification~\citep{zhang2025generative}. 
Recent evaluation and analysis work has revealed important limitations of prompted approaches, i.e., using general instruction-tuned models as verifiers. In particular, prior work has identified a positivity bias in self-verification~\citep{lu2025does, mahdavi2025scaling} and intra-family verification, as well as a tendency for weaker verifiers to overrate generations produced by stronger models~\citep{zhou2026variation, pandit2025hard2verify}. 
One line of work attempts to address these limitations by developing specialized verifiers through additional fine-tuning~\citep{zhao2025genprm, chan2025j1, ma2025s, xiong2025stepwiser}, and joint training of generators and verifiers~\citep{singh2026v_1, zha2025rl}. 
Our work investigates verification bias through the lens of hidden representations, and develops a training-free approach for improving verification behavior in general instruction-tuned models.

\textbf{Representation Engineering.} Representation engineering~\citep{zou2023representation, wu2025axbench} treats a model’s hidden representations as a central object of analysis and intervention.
Motivated by the hypothesis that features such as behaviors or concepts correspond to directions in activation space~\citep{park2023linear, nanda-etal-2023-emergent, bereska2024mechanistic, marks2024the}, prior work has used activation steering~\citep{turner2023steering} to control model behavior.
This approach has been shown to effectively steer a range of model behaviors in question-answering settings, including refusal behavior in safety-related contexts~\citep{lee2025programming, vu2025angular, xiong2026steering, sharma2026coldsteer} and hallucination reduction~\citep{wang2025semanticsadaptive, you2026spherical, zhao-etal-2025-steering}.
More recent work suggests that representation engineering is also effective for complex long-CoT reasoning.
This line of work is motivated by the hypothesis that hidden representations encode and plan future outputs beyond the next token, including longer reasoning trajectories~\citep{maar2026whats, dong2025emergent, feng2026finegrained}. 
Building on this view, prior studies have used activation-level interventions to suppress or induce thinking and reflection behaviors~\citep{chen2025seal, liu2025fractional, lin2025controlling, azizi2025activation, zhao2025activation, zhang2025understanding}, with applications to more efficient reasoning. 
Concurrent to our work, \citet{kumaran2026llms} show that verbal confidence representations emerge at answer-adjacent positions during self-evaluation. 
Our work differs in scope: we study step-wise verification where the generator and verifier are distinct models, and develop these insights into a practical method for improving verification through targeted control of verifier strictness.\looseness-1

\vspace{-2mm}
\section{Preliminaries and Background}\label{sec:preli}
\vspace{-2mm}
\textbf{Step-Wise Verification.} We consider the task of step-level error identification using a generative verifier.
Given a math problem $p$ and its step-by-step solution $\mathcal{S}=\{s_0,\ldots,s_{n-1}\}$, the verifier is prompted to predict the index of the first incorrect step in the solution.
Specifically, it outputs an index $i \in \{-1,0,\ldots,n-1\}$, where $i=-1$ indicates that all steps are correct, and $i \geq 0$ indicates that the earliest error occurs at step $s_i$.
Before producing the final verdict, the verifier generates a verification CoT $\mathcal{V}$.
We observe that the verification CoT also exhibits a step-wise structure: each verification step evaluates a single solution step, and adjacent verification steps are often separated by a double-newline token sequence (``\texttt{\textbackslash n\textbackslash n}'').
This structure is illustrated by the two examples in \Cref{fig:obs-demo}.
Accordingly, we decompose $\mathcal{V}$ as $\{v_0,\ldots,v_{n-1}\}$, where $v_i$ denotes the verification step evaluating the correctness of $s_i$.

\textbf{Evaluation Metrics.} 
Following prior work~\citep{zheng2025processbench, pandit2025hard2verify}, we evaluate step-wise verification using true positive rate (TPR), true negative rate (TNR), and their harmonic mean $\text{F1} = {2\,\mathrm{TPR}\cdot\mathrm{TNR}}/{(\mathrm{TPR}+\mathrm{TNR})}$.
TPR measures the verifier's accuracy on fully correct solutions, i.e., whether it correctly accepts all steps, while TNR measures its accuracy on incorrect solutions, i.e., whether it correctly identifies the erroneous step.\looseness-1

\textbf{Benchmarks and Datasets.}
For evaluation, we use ProcessBench~\citep{zheng2025processbench} and Hard2Verify~\citep{pandit2025hard2verify}, step-level benchmarks for identifying errors in mathematical reasoning.
ProcessBench contains four subsets spanning different difficulty levels (GSM8K, MATH, OlympiadBench, and Omni-MATH), while Hard2Verify curates challenging problems from recent international mathematics competitions (IMO, Putnam).
Both benchmarks provide human-annotated first-error locations.
For steering vector construction and correctness probe training, we use ActPRM~\citep{duan2025efficient}, a 663K-example dataset built from NuminaMath-1.5~\citep{numina_math_datasets} with step-level labels produced by an automatic pipeline combining ensemble PRMs and uncertainty estimation.
Our method uses a subsample of 11K samples from ActPRM in total.

\textbf{Categorization of Verification Mistakes.}
We categorize verification errors along a behavioral dimension we call \emph{strictness}, which describes the verifier's overall tendency to accept or reject solution steps.
When the solution contains an incorrect step $s_{i_e}$, the verifier may make an \textit{under-critical mistake}: it incorrectly concludes that all steps are correct, corresponding to a \textit{false acceptance} at step $i_e$.
The desired counterpart is a \textit{true rejection}, where the verifier correctly identifies $s_{i_e}$ as incorrect.
An example of an under-critical mistake and its correction is shown in \Cref{fig:obs-demo}.
The verifier may also predict an incorrect error location $\hat{i} \neq i_e$, which we refer to as \textit{inaccurate step identification}.
Conversely, when the solution is fully correct, the verifier may make an \textit{over-critical mistake}, incorrectly rejecting a correct step (a \textit{false rejection} at step $\hat{i}$).
The desired counterpart is a \textit{true acceptance}, where the verifier correctly judges all steps as correct.
Under this view, insufficient strictness leads to under-critical mistakes, while excessive strictness leads to over-critical behavior.

\FloatBarrier
\vspace{-2mm}
\section{Dissecting Step-Wise Verification from Latent Perspective}
\vspace{-2mm}
In \Cref{sec:begin-dissect-failure}, we present a preliminary study on the failure patterns of generative verifiers that motivates our hypothesis.
In \Cref{sec:exp-controllable}, we provide empirical evidence that verifier strictness is indeed steerable via hidden-state intervention.

\vspace{-2mm}
\subsection{Dissecting Verification Failures}\label{sec:begin-dissect-failure}
\vspace{-2mm}
We examine verification failure cases of Qwen3-8B on ProcessBench using the basic prompt (\Cref{app:prompt}), with representative examples shown in \Cref{tab:compare-veri-cot} under the column ``baseline incorrect verification''.
More examples are shown in \Cref{app:results} \Cref{tab:compare-veri-cot-complete}.

As shown in \Cref{tab:compare-veri-cot}, the baseline verifier tends to produce a superficial CoT that mainly serves to justify a pre-existing verdict: for under-critical cases, the CoT repeats erroneous intermediate steps rather than independently recalculating; for over-critical cases, it elevates superficial stylistic issues into reasoning errors. 
These patterns suggest that the verifier's accept-or-reject tendency may be determined before the corresponding verification paragraph is generated, rather than arising from step-by-step reasoning within it. 
This aligns with prior work showing that CoT in problem-solving can be ``decorative'', with the model's beliefs are already encoded in the hidden states~\citep{zhao2025can, boppana2026reasoning}.\looseness-1

Combined with the sequential structure of step-wise verification shown in \Cref{fig:obs-demo}, where adjacent verification paragraphs are separated by \textdnewline{} tokens, this motivates the following hypothesis: the verifier's tendency to accept or reject a specific solution step is encoded in the hidden state of the \textdnewline{} token preceding the corresponding verification paragraph.

\begin{table}[!ht]
\centering
\scriptsize
\setlength{\tabcolsep}{2pt}
\resizebox{0.95\linewidth}{!}{
\begin{tabular}{p{0.05\linewidth} p{0.22\linewidth} p{0.35\linewidth} p{0.35\linewidth}}
\toprule
& \multicolumn{1}{c}{\textbf{Solution Step}} &
\multicolumn{1}{c}{\textbf{Baseline Incorrect Verification}} &
\multicolumn{1}{c}{\textbf{Steered} ({\setlength{\fboxsep}{1pt}\colorbox{verylightblue}{+$\strictvwol$}}/{\setlength{\fboxsep}{1pt}\colorbox{verylightred}{+$\lenientvwol$}})} \\

\midrule
\multirow{1}{*}{\rotatebox{90}{\textbf{Under-critical}}}
& \multicolumn{1}{p{0.22\linewidth}|}{\textit{After using 25 bandages, the remaining bandages are \( \textcolor{red}{(x - 16 + 100) - 25} = \textcolor{red}{x + 69 - 25 = x + 44} \).}}
& \multicolumn{1}{p{0.35\linewidth}|}{\textbf{Analysis}: This is \textcolor{red}{correct}. They added 100 bandages, used 25, so the remaining is $ \textcolor{red}{x - 16 + 100 - 25 = x + 69 - 25 = x + 44} $. \textbf{No error}.}
& \cellcolor{verylightblue} \textbf{Analysis}: This paragraph \textcolor{red}{contains an \textbf{error}}. The expression $ \textcolor{red}{(x - 16 + 100) - 25} $ simplifies to $ \textcolor{red}{x + 84 - 25 = x + 59} $, \textcolor{red}{not $ x + 44 $}. The error is in the arithmetic: $ -16 + 100 = 84 $, not 69. So the correct remaining bandages \textcolor{red}{should be $ x + 59 $, not $ x + 44 $.} \\

\midrule

\multirow{1}{*}{\rotatebox{90}{\textbf{Over-critical}}}
& \multicolumn{1}{p{0.22\linewidth}|}{\textit{First, calculate the area of the larger rectangle. The larger rectangle has a length of 4 yards and a width of 5 yards. \([ \text{Area of the larger rectangle} = 4 \times 5 = 20 \text{ square yards} ]\)}}
& \multicolumn{1}{p{0.35\linewidth}|}{\textbf{Analysis}: ...However, the dimensions are not clearly defined in the diagram, and the interpretation of which dimension corresponds to which side is ambiguous. - \textbf{Conclusion}: \textcolor{red}{The error is in the \textbf{interpretation of the dimensions}}, ...}
& \cellcolor{verylightred}  This is also correct. The diagram labels ``4 yds'' at the top and ``5 yds'' on the left side, so the larger rectangle is 4 yards wide and 5 yards tall. The area is indeed $4 \times 5 = 20$ square yards. \\

\bottomrule
\end{tabular}}
\caption{\textbf{Representative case studies of verification failures and their correction by hidden-state steering.} Qwen3-8B on ProcessBench. \textit{Top:} an under-critical case where the baseline verifier falsely accepts an erroneous step, corrected by applying $\strictvwol$. \textit{Bottom:} an over-critical case where the baseline falsely rejects a correct step, corrected by applying $\lenientvwol$. Additional examples are provided in Appendix~\ref{app:results}.
}
\vspace{-1mm}
\label{tab:compare-veri-cot}
\end{table}

\subsection{Verification Strictness is Adjustable by Hidden-State Steering}\label{sec:exp-controllable}\vspace{-2mm}
To test the hypothesis that verifier strictness is steerable, we develop a verification-specific hidden-state steering framework with two stages: an offline stage for constructing a steering vector, and an online stage for test-time intervention. The full pipeline is shown in \Cref{fig:obs-demo} (b) and elaborated below. 
Note that this uses a standard activation steering method, contrastive activation addition (CAA)~\citep{rimsky-etal-2024-steering, wu2025axbench}.
Our goal here is not a new steering primitive but to test whether verifier strictness is steerable. 
We use CAA as a baseline and compare it with our proposed method in \Cref{sec:results}.

\textbf{Extraction of Strictness Steering Vector.} 
We first build a validation set from ActPRM~\citep{duan2025efficient}. 
To construct the steering vector for increasing verifier strictness, we use 500 randomly sampled samples labeled as erroneous.
For each sample, we use the verifier to generate 16 verification traces with temperature 0.7 and top-$p$ 0.8.
Consider an erroneous solution $\mathcal{S}$ for problem $p$, whose annotated first incorrect step is $s_{i_e}$. For each sampled verification trace $\mathcal{V}$, we segment the trace into $\{v_0,\ldots,v_{n-1}\}$ using the \textdnewline{}, and extract the verifier's final prediction $\hat{i}$. 
We retain only those erroneous samples whose 16 rollouts contain at least one trace with $\hat{i}=-1$ and at least one trace with $\hat{i}=i_e$. Recall that a verifier output of $-1$ indicates that all steps are correct.
This ensures that, for the same problem-solution pair, the verifier exhibits both false-acceptance and true-rejection behaviors across different rollouts.

We then collect traces with $\hat{i}=-1$. For erroneous solutions, such traces correspond to under-critical failures and therefore imply a false acceptance at the true erroneous step $s_{i_e}$. We then localize the corresponding verification segment $v_{i_e}$ using paragraph matching (e.g., ``Paragraph $i$'') together with acceptance-oriented keywords such as ``correct'' or ``okay''. We denote the resulting false-acceptance segment by $v_{\mathrm{FA}}$, and extract the hidden state at the delimiter token \textdnewline{} immediately preceding this segment at layer $l$, denoted by $\vh_{\mathrm{FA}}^l$. Collecting all such hidden states over the validation set gives the set $\mathcal{H}_{\mathrm{FA}}^l$.
Analogously, for traces with $\hat{i}=i_e$, we localize the true-rejection segment using rejection-oriented keywords (e.g., ``incorrect'', ``issue'') and extract the preceding delimiter-token hidden state $\vh_{\mathrm{TR}}^l$, giving the set $\mathcal{H}_{\mathrm{TR}}^l$. The complete list of keywords is provided in Appendix~\ref{app:keywords}.

We then compute the mean hidden state of each set, and define the strictness steering vector at layer $l$ as the difference between these two means:
\begin{equation}
\strictv = \bar{\vh}_{\mathrm{TR}}^l - \bar{\vh}_{\mathrm{FA}}^l,
\end{equation}
where 
$\bar{\vh}_{\mathrm{FA}}^l
=
\frac{1}{|\mathcal{H}_{\mathrm{FA}}^l|}
\sum_{\vh \in \mathcal{H}_{\mathrm{FA}}^l} \vh
$ and 
$
\bar{\vh}_{\mathrm{TR}}^l
=
\frac{1}{|\mathcal{H}_{\mathrm{TR}}^l|}
\sum_{\vh \in \mathcal{H}_{\mathrm{TR}}^l} \vh
$. 
Intuitively, $\strictv$ captures the average hidden-state direction from false acceptance to true rejection, and thus serves as a steering direction for increasing verifier strictness. For notational simplicity, we will omit the layer superscript $l$ in the following whenever it is clear from context.

Analogously, we construct the leniency steering vector using 500 fully correct ActPRM samples, collecting hidden states from true-acceptance and false-rejection segments via the same procedure:
\begin{equation}
\lenientvwol = \bar{\vh}_{\mathrm{TA}} - \bar{\vh}_{\mathrm{FR}}.
\end{equation}
This vector captures the direction from false rejection to true acceptance, and serves as a steering direction for decreasing verifier strictness.

\textbf{Test-time Intervention.} 
After obtaining the two steering directions, we test whether they can modulate verifier strictness on unseen verification samples.
Specifically, during inference, we intervene on the hidden representations of all generated \textdnewline{} tokens. Given a hidden state $\vh$ at layer $l$ and a steering direction $\vd$, we first apply a scaled perturbation and then renormalize the result to preserve the original hidden state $\ell_2$ norm:
\begin{equation}
    \vh^{\prime}
    =
    \|\vh\|_2
    \frac{\vh + \alpha \vd}{\|\vh + \alpha \vd\|_2},
\end{equation}

where $\alpha$ controls the steering strength.

\begin{figure*}[!th]
\vspace{-2mm}
\centering
    \includegraphics[width=0.95\linewidth]{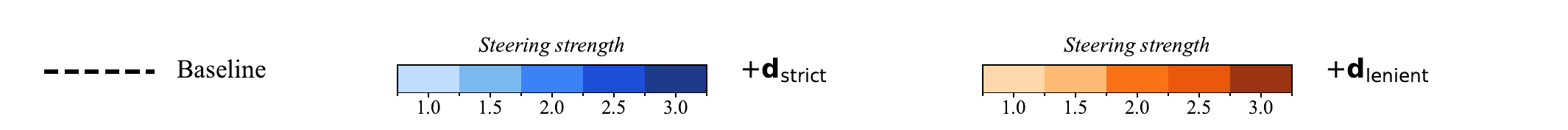} \\
    \includegraphics[width=0.95\linewidth]{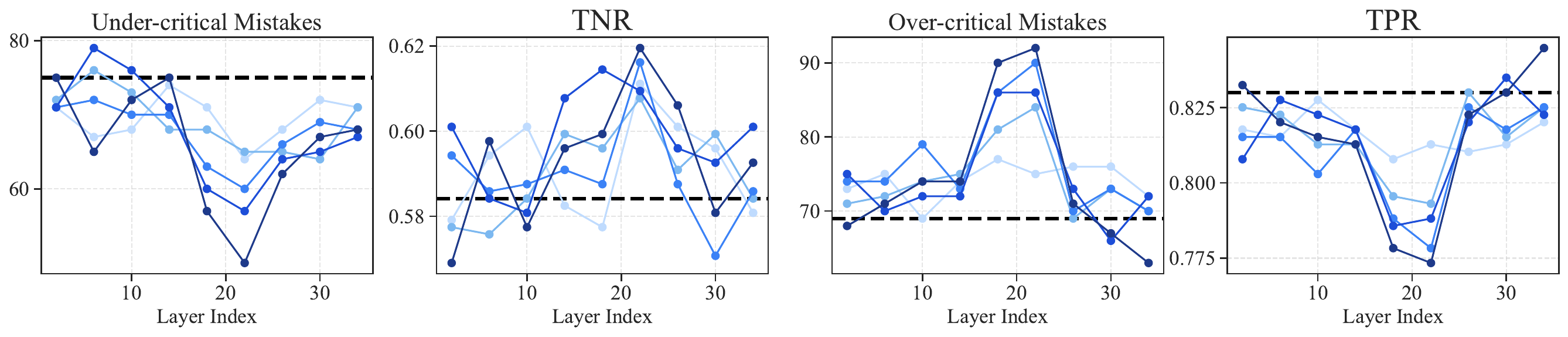}
    \includegraphics[width=0.95\linewidth]{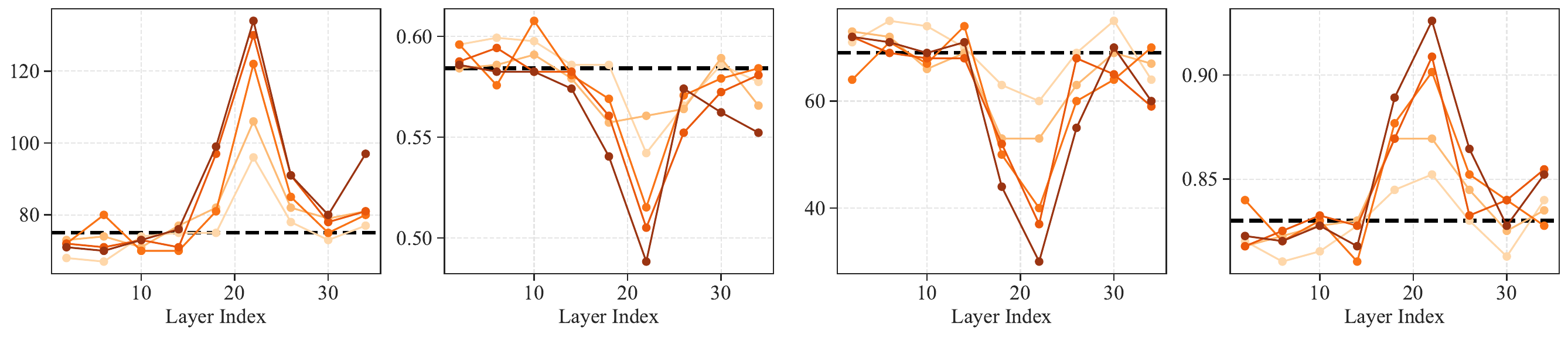}
    \vspace{-3mm}
    \caption{\textbf{Verification strictness is adjustable via hidden-state steering.}
    Results on Qwen3-8B on the Math subset of ProcessBench. 
    Each panel shows how hidden-state steering changes under-critical mistake count (lower is better), TNR (higher is better), over-critical mistake count, and TPR relative to the baseline.
    \textit{Upper}: applying the strictness-increasing vector $\strictvwol$ across different layers and steering strengths. \textit{Lower}: applying the leniency-increasing vector $\lenientvwol$. 
    \vspace{-2mm}}
\label{fig:steer_curves}
\end{figure*}

We conduct experiments on the Math subset of ProcessBench using Qwen3-8B with the standard evaluation prompt.
We sweep intervention layers every four layers across the 36-layer transformer, and vary the steering strength over $\{1.0, 1.5, 2.0, 2.5, 3.0\}$.
We use greedy decoding.
We report the results in \Cref{tab:compare-veri-cot} and \Cref{fig:steer_curves}, showing that verifier strictness is indeed adjustable through hidden-state steering.
\Cref{tab:compare-veri-cot} presents representative verification examples before and after intervention. 
We can see that steering with $\strictvwol$ converts false acceptances into true rejections, while steering with $\lenientvwol$ converts false rejections into true acceptances.
The intervention preserves fluency and semantic coherence, and does not merely flip the final verdict. 
In under-critical cases, $\strictvwol$ leads the verifier to explicitly identify the arithmetic or algebraic error and state the corrected result.
In over-critical cases, $\lenientvwol$ suppresses overly harsh judgments based on wording issues.

\Cref{fig:steer_curves} shows that hidden-state steering systematically and continuously modulates verifier strictness. The effect varies smoothly with steering strength. 
At the most effective setting, $\strictvwol$ reduces under-critical mistakes by 33.3\%, while $\lenientvwol$ reduces over-critical mistakes by 56.5\%.
As the corresponding mistake count decreases, TNR/TPR improves accordingly.
Notably, the effect concentrates around layer 22, suggesting that the model's verification decision-making is not distributed uniformly but localized to a small subset of layers that play a functionally distinct role in determining solution correctness.

\vspace{-3mm}
\section{VerifySteer: Adaptive Control of Verification Strictness}\label{sec:exp-method}
\vspace{-2mm}

\subsection{Method}\label{sec:verify_steer_method}
\vspace{-3mm}
The previous section shows that the strictness-increasing vector $\strictvwol$ and the leniency-increasing vector $\lenientvwol$ induce opposite trade-offs between TNR and TPR when applied uniformly to all samples and delimiter tokens.
This suggests that a fixed steering policy is suboptimal.
At the sample level, different verification inputs require different strictness: some benefit from increased strictness, while others require reduced strictness.
At the delimiter-token level, steering is only necessary when the local hidden state is misaligned with the desired direction; intervening on aligned states can cause inaccurate step identification.
To address this, we propose \ourmethod, a lightweight conditional steering method that first chooses the steering direction at the sample level and then applies steering selectively at the delimiter-token level.

\textbf{Sample-level Conditional Steering Policy.}
Given a verification prompt containing the math problem $p$ and candidate solution $\mathcal{S}$, we extract a prompt-level representation $\vh_p$ at layer $l$ by pooling the hidden states of the input tokens.
Following \citet{li2026rethinking}, we consider two pooling strategies: last-token and mean pooling, see Appendix~\ref{app:method} for details.
We feed $\vh_{p}$ into a lightweight probe $f_\phi$, which outputs a scalar correctness score $q(p,\mathcal{S})=\sigma(f_\phi(\vh_{p})) \in [0,1]$, where larger values indicate that the candidate solution is more likely to be correct.
We introduce two variants: \ourmethoduni (unidirectional), which applies $\strictvwol$ when the solution is predicted to be likely incorrect ($q \leq \tau_l$) and does not steer otherwise; and \ourmethodbi (bidirectional), which additionally applies $\lenientvwol$ when the solution is predicted to be likely correct ($q \geq \tau_h$), with no steering in the intermediate uncertainty region.
Formally, the selected steering direction is:
\vspace{-1mm}
\begin{equation}\label{eq:steer-dir}
\vd_{\mathrm{sel}}(p,\mathcal{S})=
\begin{cases}
\strictvwol, & q(p,\mathcal{S}) \leq \tau_l,\\
\lenientvwol, & q(p,\mathcal{S}) \geq \tau_h \text{ and } \mathsf{A}=\ourmethodbi,\\
\mathbf{0}, & \text{otherwise},
\end{cases}
\vspace{-1mm}
\end{equation}
where $\tau_l < \tau_h$ and $\mathsf{A}$ denotes the \ourmethod{} variant.

\textbf{Delimiter-level Intervention Gate.}
After selecting the steering direction $\vd_{\mathrm{sel}}$ for a sample, we further determine which delimiter tokens should be intervened on.
If $\vd_{\mathrm{sel}}=\mathbf{0}$, we leave all delimiter-token hidden states unchanged.
Otherwise, steering is only applied when the local hidden state is sufficiently misaligned with the selected steering direction.
For each delimiter token hidden state at layer $l$ $\vh_k$, we apply steering as:
\begin{equation}\label{eq:deli-gate}
\vh_k^{\prime}
=
(1-g_k)\,\vh_k
+
g_k \,
\|\vh_k\|_2
\frac{\vh_k+\alpha \vd_{\mathrm{sel}}}{\|\vh_k+\alpha \vd_{\mathrm{sel}}\|_2},
\end{equation}
where
$
g_k = \mathbf{1}\!\left[\text{cossim}(\vh_k,\vd_{\mathrm{sel}})<\rho\right]
$
is a binary gate, $\text{cossim}$ denotes the cosine similarity.
This gate prevents unnecessary intervention when the local hidden state is already sufficiently aligned with the selected steering direction and therefore does not require correction.

\vspace{-3mm}
\subsection{Results}\label{sec:results} 
\vspace{-3mm}

\textbf{Experimental Setup.} We compare \ourmethod{} against baselines on all ProcessBench subsets and Hard2Verify. 
For ProcessBench, we use Qwen3-1.7B and Qwen3-8B~\citep{yang2025qwen3} in non-thinking mode. For Hard2Verify, we evaluate two verifiers: Qwen3-8B in thinking mode (denoted Qwen3-8B-thinking), and FARE-20B~\citep{xu2026foundational}, a multi-task evaluator fine-tuned from the MoE model gpt-oss-20B~\citep{openai2025gptoss120bgptoss20bmodel} with 2.5M training samples.
We also report results of large instruction-tuned models Qwen3-14B, Llama-3.3-70B, Qwen2.5-72B and gpt-oss-20B (low reasoning) for reference.
We report TNR, TPR, and F1 for verification performance, and TFLOPs per sample for inference cost. 
Following \citet{snell2024scaling}, we approximate inference FLOPs as $2MT_{\text{inf}}$, where $M$ is the number of model parameters and $T_{\text{inf}}$ is the number of inference tokens. For the MoE model gpt-oss-20B and FARE-20B, $M$ refers to the number of active parameters per token.\looseness-1

We consider three baselines: (i) a \textit{base evaluation prompt}, using the standard ProcessBench prompt~\citep{zheng2025processbench}; (ii) an \textit{optimized prompt}, adapted from \citet{zhang2025lessons} for ProcessBench and from \citet{pandit2025hard2verify} for Hard2Verify, which instructs the model to be more critical; and (iii) \textit{self-consistency} ($\texttt{Cons@N}$), which generates $N$ verification traces via sampling-based decoding and selects the final verdict by majority voting. 
The inference cost of self-consistency is approximately $N$ times more than other methods.
For the fine-tuned verifier FARE-20B, we use the prompt it was fine-tuned with, denoted as the \textit{FARE Prompt}.
All prompts are provided in \Cref{app:prompt}.
We find that the optimized prompts are benchmark-specific and do not transfer well across benchmarks.
For our method \ourmethod, we use the optimized prompt for each benchmark. 
We construct steering vectors using validation samples from ActPRM, and train a three-layer MLP probe $f_\phi$ on validation samples.
Details on validation set and hyperparameters are provided in Appendix~\ref{app:method}.\looseness-1

\begin{table}[!th]
\centering 
\renewcommand{\arraystretch}{1.25} 
\resizebox{\textwidth}{!}{
\begin{tabular}{l ccc ccc ccc ccc c@{\hskip 6pt}c}
\toprule
& \multicolumn{3}{c}{\textbf{GSM8K}} & \multicolumn{3}{c}{\textbf{Math}} & \multicolumn{3}{c}{\textbf{OlympiadBench}} & \multicolumn{3}{c}{\textbf{Omni-MATH}} & \multicolumn{2}{c}{\textbf{Overall}} \\
\cmidrule(lr){2-4} \cmidrule(lr){5-7} \cmidrule(lr){8-10}  \cmidrule(lr){11-13} \cmidrule(lr){14-15}
\textbf{Metric} & \textbf{TNR} & \textbf{TPR} & \textbf{F1} & \textbf{TNR} & \textbf{TPR} & \textbf{F1} & \textbf{TNR} & \textbf{TPR} & \textbf{F1} & \textbf{TNR} & \textbf{TPR} & \textbf{F1} & \textbf{F1} & \textbf{TFLOPs} \\
\midrule
\multicolumn{14}{c}{\textit{Qwen3-1.7B}} \\
Basic Prompt          & 17.9 & 96.4 & 30.2 & 14.5 & 84.2 & 24.7 & 7.7 & 78.8 & 14.1 & 8.0 & 82.6 & 14.7 & 20.9 & 2.86\\
Prompt Opt.          & 22.2 & 76.7 & 34.5 & 18.0 & 58.4 & 27.5 & 13.3 & 49.6 & 21.0 & 10.4 & 57.3 & 17.6 & 25.1 & 3.19 \\
\hspace{4mm}  +\texttt{Cons@4}$^\dag$  &  21.6 &  81.2 &  34.1  & 20.3  & 69.4 & 31.4 & 11.0 & 58.5 & 18.5 & 12.0 & 62.2 & 20.0 & 26.0 & 12.71 \\
\rowcolor{LightBlue} \ourmethoduni & 25.6 & 77.2  & \underline{38.5}  & 21.7 &  61.6  &  \underline{32.1} &  15.6  &  56.6  &  \textbf{24.4} & 12.6 & 59.3 & \underline{20.9} & \underline{29.0}  & 3.37\\
\rowcolor{LightBlue} \ourmethodbi & 27.1 & 83.4  & \textbf{40.9}  & 21.5 &  69.2  &  \textbf{32.9} &  15.0  &  58.4  &  \underline{23.8} & 12.6 & 64.3 & \textbf{21.1} & \textbf{29.7} & 3.47 \\
\midrule
\multicolumn{14}{c}{\textit{Qwen3-8B}} \\
Basic Prompt          & 54.6 & 96.4 & 69.7 & 58.4 & 83.0 & 68.6 & 54.0 & 76.1 & 63.2 & 53.8 & 73.4 & 62.1 & 65.9 & 16.71 \\
Prompt Opt. & 58.5 & 94.3 & 72.2 & 66.8 & 85.0 & 74.8 & 58.4 & 77.9 & 66.7 & 58.6 & 75.5 & 66.0 & 69.9 & 21.26 \\
\hspace{4mm}  +\texttt{Cons@4}$^\dag$ & 63.3 & 95.6 & \underline{76.2} &  69.5 & 86.3 & \textbf{77.0} & 61.3 & 77.9 & \textbf{68.7} & 63.1 & 71.4 & 67.0 & \underline{72.2} & 84.09 \\

\rowcolor{LightBlue} \ourmethoduni  & 62.3 & 93.3 & 74.7 & 69.0 & 85.5 & 76.4 & 59.6 & 77.0 & 67.2 & 61.9 & 76.8 & \underline{68.5} & 71.7 & 21.23 \\
\rowcolor{LightBlue} \ourmethodbi  &  64.3 & 93.8 & \textbf{76.3} & 68.7 & 86.9 & \underline{76.7} & 60.1 & 76.7 & \underline{67.4} & 61.8 & 77.2 & \textbf{68.6} & \textbf{72.3} & 21.23 \\
\midrule
\textit{Qwen3-14B}     &  61.4 & 98.4 & 75.6 & 68.5 & 86.9 & 76.6 & 58.9 & 74.3 & 65.7 & 59.6 & 80.2 & 68.4 & 71.6 & 37.45 \\
\textit{Llama-3.3-70B} & 68.6 & 96.9 & 80.3 & 46.0 & 94.3 & 61.8 & 32.8 & 93.2 & 48.6 & 29.0 & 88.0 & 43.6 & 58.6 & 85.45 \\
\textit{Qwen2.5-72B} & 61.8 & 95.9 & 75.2 &  45.8 & 91.4 & 61.0 & 35.9 & 87.6 & 50.9 & 37.5 & 83.4 & 51.8 & 59.7 & 53.03 \\
\bottomrule
\end{tabular}} 

\caption{Comparison of \ourmethod with baselines on ProcessBench.
\texttt{Cons@4} refers to self-consistency method with 4 sampled results and the final answer is selected by majority voting.
$^\dag$ denotes sampling-based decoding with temperature 0.7 and top-$p$ 0.8.
We report the mean across 4 runs of \texttt{Cons@4}.
Methods without $^\dag$ use greedy decoding. TFLOPs reports the estimated inference compute per sample.
Best and second-best results are \textbf{bolded} and \underline{underlined} within each Qwen3-1.7B/8B block and benchmark subset.
}
\label{tab:verify_steer}
\end{table}

\textbf{Main Results.} 
The results are shown in \Cref{tab:verify_steer,tab:verify_steer_h2v}.
Compared with the base prompt and prompt optimization baselines, both \ourmethod{} variants outperform these baselines across all models and benchmarks, showing that hidden-state steering provides additive gains on top of prompt-level optimization, and it is compatible with different prompts.
\ourmethod{} closes the performance gap to larger models.
With \ourmethod{}, Qwen3-8B surpasses Qwen3-14B on ProcessBench (72.3 vs.~71.6 overall F1), and Qwen3-8B-thinking surpasses Qwen3-14B-thinking on Hard2Verify (33.4 vs.~30.4); neither holds without our method.\looseness-1

Compared with self-consistency, on ProcessBench \ourmethod{} surpasses \texttt{Cons@4} in F1 on Qwen3-1.7B by a margin of 3.7 and achieves comparable performance on Qwen3-8B.
We can see \texttt{Cons@4} requires around $4\times$ the inference FLOPs of our method and other baselines.
On Hard2Verify, \ourmethod{} surpasses \texttt{Cons@8} by a margin of 2.2 F1 while requiring $7\times$ fewer FLOPs.
A closer look reveals that the self-consistency baseline does not effectively improve TNR, suggesting that basic parallel test-time scaling is less effective at mitigating evaluator positivity bias compared to hidden-state steering.

\begin{wraptable}{r}{0.48\textwidth}
\vspace{-4mm}
\centering
\renewcommand{\arraystretch}{1.25}
\resizebox{0.45\textwidth}{!}{
\begin{tabular}{lccc@{\hskip 6pt}c}
\toprule
\textbf{Method} & \textbf{TNR} & \textbf{TPR} & \textbf{F1}\footnotemark & \textbf{TFLOPs} \\
\midrule
\multicolumn{5}{c}{\textit{Qwen3-8B-thinking}} \\
Basic Prompt              & 12.8 & 94.0 & 22.4 & 72.64 \\
Prompt Opt.              & 18.0 & 78.6 & 29.1 & 69.59 \\
\hspace{5mm} +\texttt{Cons@4}     & 18.7 & 88.9 & 30.7 & 274.33 \\
\hspace{5mm} +\texttt{Cons@8}     & 18.8 & 92.1 & 31.2 & 545.19 \\
\rowcolor{LightBlue} \ourmethoduni & 21.5 & 75.6 & \underline{33.4} & 77.72 \\
\rowcolor{LightBlue} \ourmethodbi  & 21.5 & 76.2 & \textbf{33.4} & 77.71 \\
\midrule
\multicolumn{5}{c}{\textit{FARE-20B}} \\
FARE Prompt & 28.5 & 76.2 & 41.5 & 7.7 \\
\hspace{5mm} +\texttt{Cons@4} &27.4 &  85.7 & 41.5 & 30.9    \\
\hspace{5mm} +\texttt{Cons@8} & 28.3 & 87.3 &  42.7 &  62.1 \\
\rowcolor{LightBlue} \ourmethoduni & 35.4 & 73.8 & \underline{47.9} & 9.2   \\
\rowcolor{LightBlue} \ourmethodbi  & 37.3 & 81.0 & \textbf{51.1} & 9.9 \\
\midrule
\textit{gpt-oss-20B} & 19.0 & 90.5 & 31.4 & 5.0 \\
\textit{Qwen3-14B-thinking} & 18.5 & 85.4 & 30.4 & 110.87\\
\textit{Llama-3.3-70B} & 4.4 & 88.1 & 8.4 & 71.35\\
\textit{Qwen2.5-72B} & 10.1 & 69.0 & 17.7 & 53.03 \\ 
\bottomrule
\end{tabular}
}
\caption{Comparison of \ourmethod with baselines on Hard2Verify. For Qwen3-8B-thinking and Qwen3-14B-thinking, we use the recommended sampling settings (temperature 0.6 and top-$p$ 0.95) and report the mean over 8 runs.
\texttt{Cons@4} and \texttt{Cons@8} are averaged over 6 self-consistency runs.  TFLOPs reports the estimated inference compute per sample. 
Best and second-best results are \textbf{bolded} and \underline{underlined} within each main model block.
\looseness-1 
}
\vspace{-2mm}\label{tab:verify_steer_h2v}
\end{wraptable}

\footnotetext{TNR, TPR, and F1 are each averaged across runs. Because the harmonic mean is a nonlinear function, the averaged F1 may not exactly equal the harmonic mean of the averaged TNR and TPR.}

\ourmethod{} is orthogonal to and compatible with verification fine-tuning. 
In \Cref{tab:verify_steer_h2v}, fine-tuning gpt-oss-20B into FARE-20B raises F1 from 31.4 to 41.5, and applying \ourmethod{} on top further raises F1 to 51.1, indicating that the gains from steering and fine-tuning stack rather than overlap. 
\ourmethod{} can therefore serve as a drop-in addition on top of fine-tuned generative verifiers. 
We also observe that, on ProcessBench, \ourmethod{} improves Qwen3-1.7B from 25.1 to 29.7 in overall F1, surpassing several fine-tuned PRMs based on earlier-generation models: Qwen2.5-Math-7B-Math-Shepherd (28.9), RLHFlow-PRM-Mistral-8B (28.4)~\citep{xiong2024implementation}, and RLHFlow-PRM-Deepseek-8B (26.6), as reported in \citet{zhang2025lessons}.
These fine-tuned PRMs required substantial training resources (e.g., 445K samples for Qwen2.5-Math-7B-Math-Shepherd) and are 4--5$\times$ larger than the steered model.
This suggests that lightweight steering applied to a current-generation model can be competitive with specialized fine-tuned evaluators built on older-generation models.

The two variants also exhibit behavior consistent with their intended roles.
\ourmethoduni, which uses only the strictness-increasing vector $\strictvwol$, generally improves TNR while largely preserving TPR unchanged.
\ourmethodbi, which additionally uses the leniency-increasing vector $\lenientvwol$, usually improves both TNR and TPR.\looseness-1

\begin{figure*}[!th]
\vspace{-3mm}
\centering
    \includegraphics[width=0.95\linewidth]{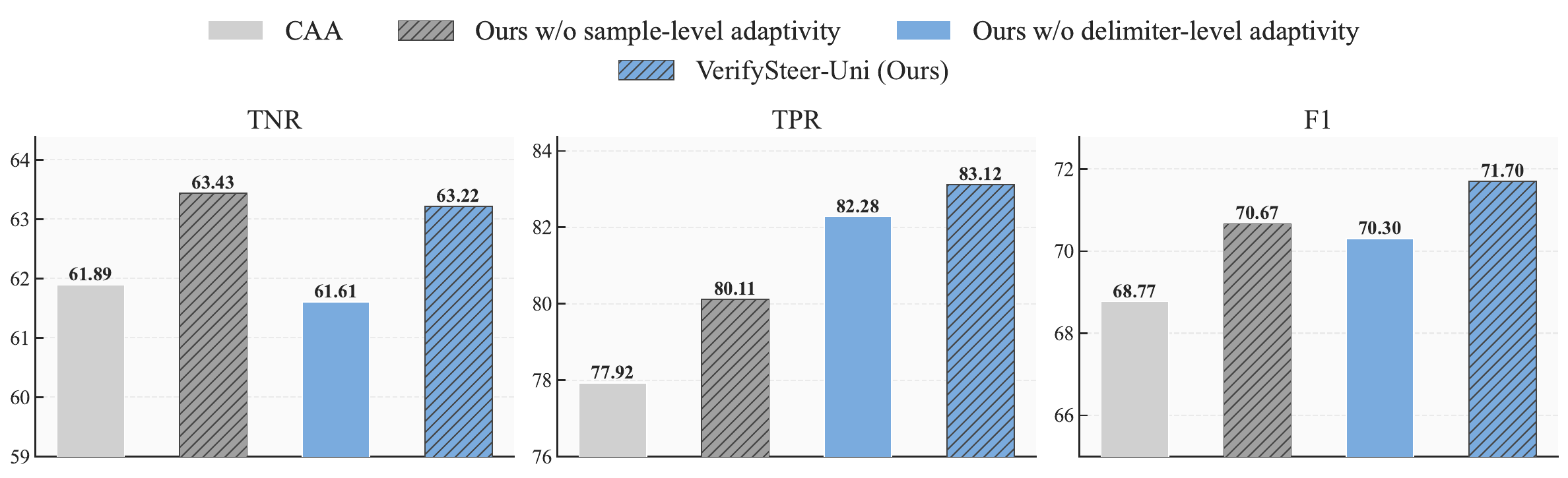}
    \caption{\textbf{Additional baseline comparisons and ablations.}
    From left to right, the three subplots report TNR, TPR, and F1, averaged over the four ProcessBench subsets.
    The leftmost gray bar corresponds to the activation steering baseline CAA~\citep{rimsky-etal-2024-steering}.
    The gray-hatched and blue bars show ablations of \ourmethod without sample-level adaptivity and without delimiter-level adaptivity.
    The rightmost bar shows the full method. We use Qwen3-8B with greedy decoding. 
    }
    \label{fig:baseline-ablation}
    \vspace{-1mm}
\end{figure*}

\textbf{Additional Baseline Comparisons and Ablations.}
We further compare our method against CAA~\citep{rimsky-etal-2024-steering}, an activation steering baseline.
Both methods only use the same strictness-increasing direction $\strictvwol$.
CAA also uses the same steering vector construction and validation set as described in \Cref{sec:exp-controllable}.
In addition, we ablate the two adaptive components of our method introduced in \Cref{sec:exp-method}.
The results are summarized in \Cref{fig:baseline-ablation}.

CAA achieves a TNR similar to \ourmethoduni, but at the cost of a large reduction in TPR, which lowers F1.
This indicates that uniform steering can improve strictness, but often does so too aggressively.
\ourmethoduni\ preserves both high TNR and high TPR, showing the effect of adaptive steering.
The ablation results further highlight the importance of each component.
Sample-level adaptivity is important for balancing TNR and TPR, suggesting that input-dependent control of strictness is beneficial.
Removing delimiter-level adaptivity lowers TNR. 
This is likely because, without delimiter-level adaptivity, steering becomes less precisely targeted to the relevant verification steps.

\section{Conclusion}
We study step-wise verification through the lens of hidden-state steering.
Our central finding is that verifier strictness is encoded in a verification-specific latent signal that can be modulated without fine-tuning.
While simple uniform steering improves either error detection or correctness certification, it induces an inherent trade-off.
\ourmethod{} addresses this through sample-level routing and selective paragraph-level intervention.
Experiments on ProcessBench and Hard2Verify show that our method outperforms prompting and self-consistency baselines.
Notably, \ourmethod{} matches or surpasses self-consistency while requiring $4$--$7\times$ less inference compute.
Our results suggest that LLM verifiers reflect controllable latent behaviors, opening a path toward lightweight and adaptive control of verifier behavior in general instruction-tuned models.
\looseness-1
\section*{Acknowledgement}

The work was supported by Department of Defense under grant HT9425-23-1-0267. Our conclusions do not necessarily reflect the position or the policy of our sponsors, and no official endorsement should be inferred.

\newpage

\bibliography{colm2026_conference}
\bibliographystyle{colm2026_conference}

\appendix
\crefalias{section}{appendix}
\newpage

\section{Prompt Templates}\label{app:prompt}
We use three prompt templates for step-wise verification evaluation, described below.

\textbf{Basic Prompt.} This is the standard evaluation template provided by \citet{zheng2025processbench}, which instructs the verifier to review the solution paragraph by paragraph and return the index of the first erroneous step.

\begin{tcolorbox}[breakable,enhanced, left=-1cm, right=-1cm, top=2pt, bottom=2pt, enlarge top by=0.1cm, enlarge bottom by=0.1cm, title={\hspace{1cm} Basic Prompt}, fonttitle=\bfseries\small]
\begin{quote}
\begin{lstlisting}
### User Prompt
The following is a math problem and a solution (split into paragraphs, enclosed with tags and indexed from 0):

[Math Problem]

{problem}

[Solution]

{tagged_response}

Your task is to review and critique the solution paragraph by paragraph. Once you identify an error in a paragraph, return the index of the paragraph where the earliest error occurs. Otherwise, return the index of -1 (which typically denotes ``not found'').

Please put your final answer (i.e., the index) in \\boxed{{}}.
\end{lstlisting}
\end{quote}
\end{tcolorbox}

\textbf{Optimized Prompt for ProcessBench.}
This prompt is adapted from \citet{zhang2025lessons}, which was originally used by LLM-as-judge to construct PRM training data. 
It extends the basic prompt with instructions for more detailed analysis, including explicit recalculation and reflection. 
It also instructs the model to be ``as critical as possible'', aiming to increase verifier strictness and improve error detection. 
This prompt achieves F1 gains across all models and subsets on ProcessBench.

\begin{tcolorbox}[breakable,enhanced, left=-1cm, right=-1cm, top=2pt, bottom=2pt, enlarge top by=0.1cm, enlarge bottom by=0.1cm, title={\hspace{1cm} Optimized Prompt for ProcessBench}, fonttitle=\bfseries\small]
\begin{quote}
\begin{lstlisting}
### User Prompt
The following is a math problem and a solution (split into paragraphs, enclosed with tags and indexed from 0):

[Math Problem]

{problem}

[Solution]

{tagged_response}

Your task is to review and critique the solution paragraph by paragraph. Once you identify an error in a paragraph, return the index of the paragraph where the earliest error occurs. Otherwise, return the index of -1 (which typically denotes ``not found'').

Review each paragraph carefully for mathematical correctness and logical consistency. Be as critical as possible.
When you analyze each paragraph, you should use proper verification, recalculation, or reflection to indicate whether it is logically and mathematically valid. Please elaborate on the analysis process carefully.

Please put your final answer (i.e., the index) in \\boxed{{}}.
\end{lstlisting}
\end{quote}
\end{tcolorbox}

\textbf{Optimized Prompt for Hard2Verify.}
The ProcessBench optimized prompt does not transfer well to Hard2Verify, where it fails to improve the thinking model Qwen3-8B-thinking.
We therefore use a separate prompt adapted from \citet{pandit2025hard2verify}, which includes a system prompt instructing the model to act as a ``strict, reliable math grader'' and provides more structured instructions for checking mathematical correctness and logical consistency.

\begin{tcolorbox}[breakable,enhanced, left=-1cm, right=-1cm, top=2pt, bottom=2pt, enlarge top by=0.1cm, enlarge bottom by=0.1cm, title={\hspace{1cm} Optimized Prompt for Hard2Verify}, fonttitle=\bfseries\small]
\begin{quote}
\begin{lstlisting}
### System Prompt
You are a strict, reliable math grader performing an Error identification task.
Your task is to identify the first incorrect paragraph in a mathematical solution.
Return your answer using the exact format requested by the user.

### User Prompt
The following is a math problem and a solution (split into paragraphs, enclosed with tags and indexed from 0):

[Math Problem]

{problem}

[Solution]

{tagged_response}

Your task is to identify the first incorrect paragraph in the solution.

Instructions:
- Review each paragraph carefully for mathematical correctness and logical consistency
- A paragraph is incorrect if it contains mathematical errors, logical inconsistencies, or is based on errors from previous paragraphs
- Find the FIRST paragraph that is incorrect (0-indexed)
- If ALL paragraphs are correct, return -1

Provide your reasoning and then put your final answer (i.e., the index) in \\boxed{{}}.
\end{lstlisting}
\end{quote}
\end{tcolorbox}

\textbf{FARE Prompt.} This is the prompt used by \citet{xu2026foundational} to fine-tune the FARE-series evaluator. 
We use this prompt for FARE-20B to ensure consistency with its fine-tuning setup.

\begin{tcolorbox}[breakable,enhanced, left=-1cm, right=-1cm, top=2pt, bottom=2pt, enlarge top by=0.1cm, enlarge bottom by=0.1cm, title={\hspace{1cm} FARE Prompt}, fonttitle=\bfseries\small]
\begin{quote}
\begin{lstlisting}
### System Prompt
Please act as an impartial judge and evaluate the quality of the response provided by an AI assistant to the user prompt displayed below. You will be given the assistant's solution to a math problem, which is split into steps, starting with a <step [step number]> tag, where [step number] is indexed from 0. Your job is to identify which step an error occurs, if an error is present.
When evaluating the solution, consider each step separately. Evaluate the content of each step for correctness. If you encounter a mistake at <step [step number]>, output [step number] as your Verdict. If the full response is error free, then select step number -1. Avoid any biases, such as length of step, or stylistic elements like formatting.

Here are some rules for evaluation.
(1) The assistant's answer does not need to be complete or arrive at a final solution. You may receive a partially complete response. Your job is to assess the quality of each step.
(2) When evaluating the assistant's answer, identify any mistakes or inaccurate information. Focus on the content each step and determine if the step is logically valid.
(3) For each step, you should provide an explanation of your assessment. If you find an error, describe the nature and cause of the error.
(4) Avoid any biases, such as answer length, or stylistic elements like formatting.

Before providing an your final verdict, think through the judging process and output your thoughts as an explanation
After providing your explanation, you must output the corresponding step number with an error. Use the following format:
Explanation: Your explanation here
Verdict: The step number with the error or -1 if no error occurs

### User Prompt
[User Question]
{instruction}

[The Start of Assistant's Answer]
{response}
[The End of Assistant's Answer]
\end{lstlisting}
\end{quote}
\end{tcolorbox}

\section{Implementation and Hyperparameters}\label{app:imple-hyper}

\subsection{Keywords for Identifying Verification Paragraphs}\label{app:keywords}

As described in \Cref{sec:exp-controllable}, we use a simple keyword-based heuristic to identify acceptance and rejection verification paragraphs.
We first localize the relevant candidate paragraph using the step index, either from the ground-truth first-error annotation or from the verifier's predicted first error, together with the paragraph marker ``paragraph $i$''.
We then use keyword matching as a consistency check to confirm that the localized paragraph contains the expected verdict cue and to filter out paragraphs with conflicting cues.
Specifically, an acceptance paragraph (either false-acceptance or true-acceptance) must contain at least one required acceptance cue and none of the excluded cues; false-rejection and true-rejection paragraphs are identified analogously.
We additionally exclude paragraphs containing ``let's'' or ``let me'', which frequently appear in thinking-model transition sentences (e.g., ``let me check if this is correct'') without conveying an actual verification verdict.
The full keyword list is shown in \Cref{tab:keywords}.

\begin{table}[!h]
    \centering
    \begin{tabular}{m{2cm}|m{8cm}}  
        \toprule
        \multirow{2}{*}{Acceptance} 
        & \textbf{Required cues}: correct, okay, no error \\  
        \cmidrule{2-2}
        & \textbf{Excluded cues}: incorrect, the correct, not correct, **not** correct, let me, let's \\  
        \midrule
        \multirow{2}{*}{Rejection}
        & \textbf{Required cues}: error, incorrect, issue, mistake, flaw, inconsistency, not correct, wrong \\  
        \cmidrule{2-2}
        & \textbf{Excluded cues}: no/any error, any explicit/immediate/mathematical error, no immediate/mathematical error, is logically/mathematically correct, not a mathematical error, does not contain an error, \, correct, let me, let's \\  
        \bottomrule
    \end{tabular}
    \caption{Keyword-based rules used to identify acceptance and rejection verification paragraphs. A paragraph is labeled as Acceptance if it contains at least one required acceptance cue and none of the excluded cues; the same rule is applied analogously for Rejection.}
    \label{tab:keywords}
\end{table}

\subsection{Steering Method}\label{app:method}

Here we provide implementation details for \ourmethod.

\textbf{Validation set construction.}
We construct validation sets from ActPRM for both steering vector extraction and correctness probe training. 
ActPRM is built on problems from NuminaMath-1.5~\citep{numina_math_datasets}, which provides source labels for each problem. 
We exploit these labels to sample subsets that are aligned with ProcessBench subsets in terms of problem difficulty and type, using fuzzy matching on problem text to identify source-aligned examples. 
Specifically, for the GSM8K subset, we randomly sample 1,000 math word problems from ActPRM. 
For the MATH subset, we collect 4,224 examples whose source problems are matched to the MATH training set~\citep{dan_henrycks_math} and MetaMath~\citep{yu2024metamath}. 
For OlympiadBench and Omni-MATH, we sample 5,132 olympiad-level examples. 
For Hard2Verify, we sample 812 competition-level proof problems using keyword matching of competition names (e.g., IMO, Putnam, USAMO) following the problem categorization in \citet{pandit2025hard2verify}. 
For each sample, we generate 16 verification rollouts using the verifier model to be steered with the optimized prompt.

\textbf{Correctness probe training.}
The correctness probe $f_\phi$ is a three-layer MLP with ReLU activations and dropout 0.1.
We train with cross-entropy loss using Adam with learning rate $10^{-5}$, cosine learning-rate scheduler, weight decay $10^{-2}$, batch size 32, and a maximum of 300 epochs.
These hyperparameters are shared across all settings.
The probe input layer is selected based on validation AUC: layer 23 for Qwen3-8B and Qwen3-8B-thinking, and layer 17 for Qwen3-1.7B.
For pooling, we use mean pooling for Qwen3-8B and Qwen3-8B-thinking, and last-token pooling for Qwen3-1.7B.
For ProcessBench, we train on 7,152 ActPRM samples with a held-out validation set of 4,768 samples.
For Hard2Verify, we train on 2,448 samples with 1,050 validation samples.

\begin{figure*}[!th] 
\centering
    \begin{subfigure}{0.32\linewidth}
    \includegraphics[width=\linewidth]{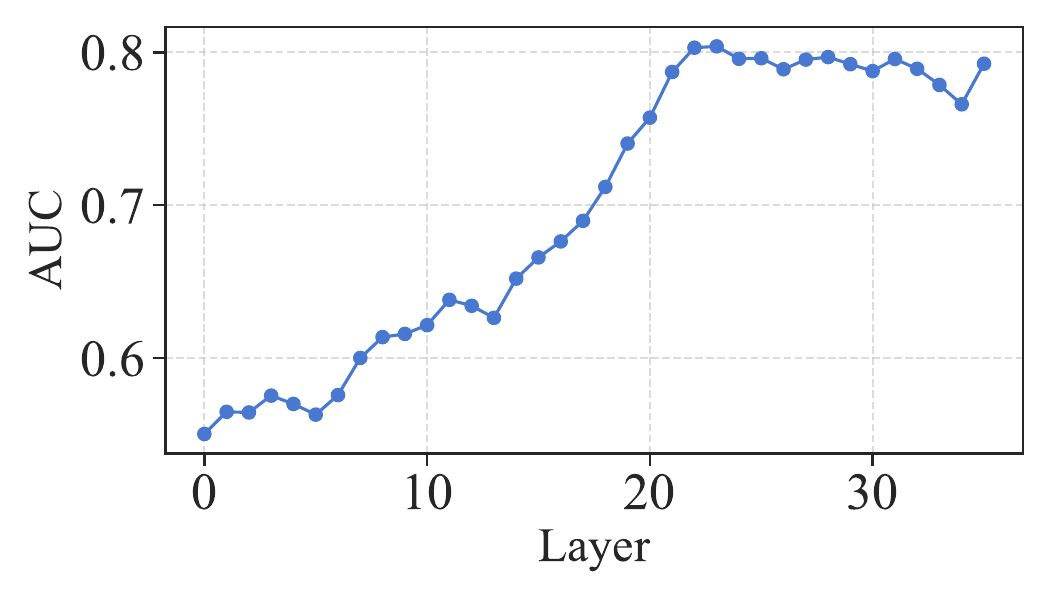}
    \caption{Qwen3-8B} 
    \end{subfigure}
    \begin{subfigure}{0.32\linewidth}
    \includegraphics[width=\linewidth]{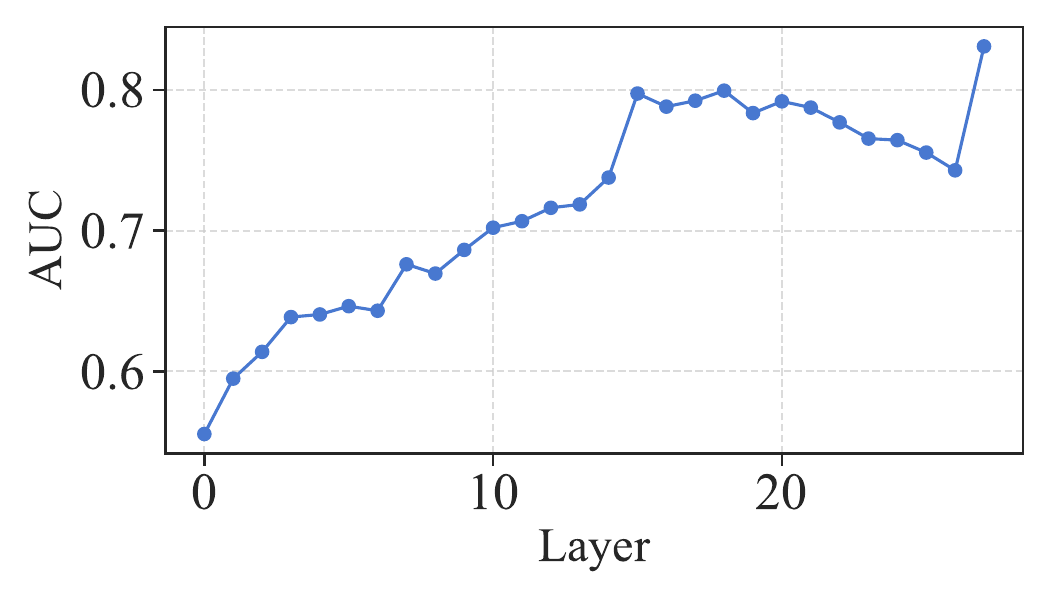} 
    \caption{Qwen3-1.7B}  
    \end{subfigure}
    \begin{subfigure}{0.32\linewidth}
    \includegraphics[width=\linewidth]{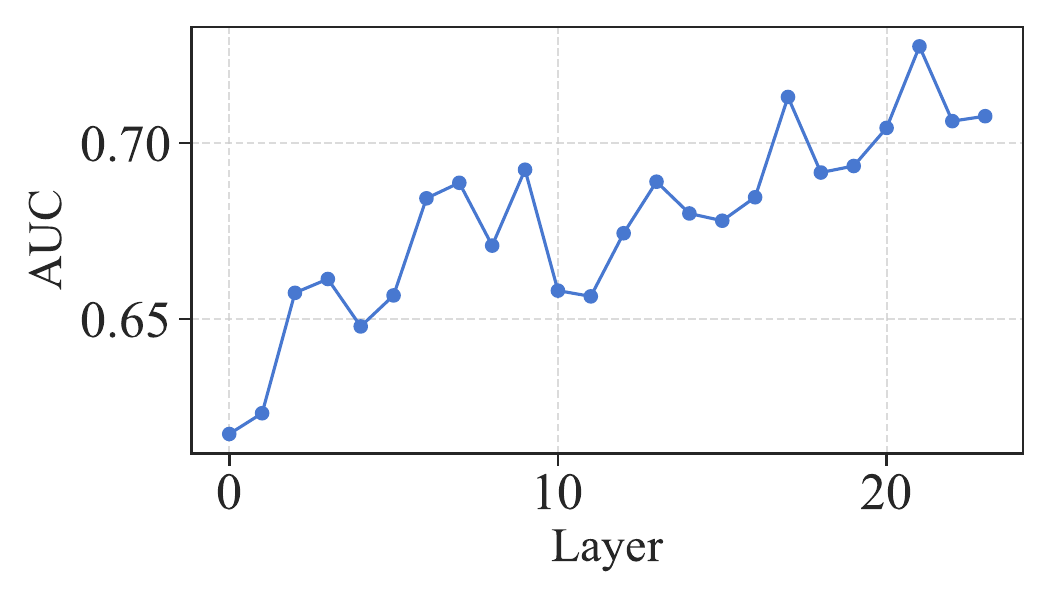} 
    \caption{FARE-20B}  
    \end{subfigure}
    \caption{Validation AUC of linear classifier on classifying delimiter tokens before the true rejection and false acceptance paragraphs.}\label{fig:auc-layer}
\end{figure*}

\textbf{Steering hyperparameters.}
For steering layer selection, we follow the criterion and procedure used in prior work. The criterion is based on the linear separability~\citep{park2023linear} of LLM concepts and behaviors in hidden states: \citet{li2025fairsteer, zhang2025understanding, lin2025controlling} identify the steering layer as the one where the target features are most separable.

Concretely, we sample 1{,}000 problems from ActPRM, generate 16 verification rollouts per problem, and collect delimiter-token hidden states preceding true rejection and false acceptance paragraphs (identified using the keywords in \Cref{app:keywords}) to form the training set. 
We repeat the same collection on a disjoint set of 1{,}000 problems to form the validation set. 
Following \citet{li2025fairsteer, lin2025controlling}, for each layer we fit a linear classifier to distinguish true rejection from false acceptance hidden states and report the validation AUC. 
A higher AUC indicates that the two verification patterns are more separable at that layer. 
\Cref{fig:auc-layer} shows the AUC curves across layers for the three models we study. 
Following \citet{lin2025controlling}, for each model we first shortlist the top-6 layers by validation AUC, and then determine the two final steering layers based on early steering performance on the smallest ProcessBench subset, GSM8K.

\Cref{tab:hyperparams} summarizes the steering hyperparameters for each model and benchmark setting.
For all settings, \ourmethoduni{} uses only $\strictvwol$ with routing threshold $\tau_l$, while \ourmethodbi{} additionally applies $\lenientvwol$ with routing thresholds $(\tau_l, \tau_h)$.
$\rho$ is the delimiter-level cosine-similarity threshold introduced in \Cref{eq:deli-gate}.

\begin{table}[!h]
\centering
\label{tab:hyperparams}
\resizebox{\columnwidth}{!}{%
\begin{tabular}{lccccccccc}
\toprule
Benchmark & Model & Variant & Layers & $\alpha_{\text{strict}}$ & $\alpha_{\text{lenient}}$ & $\tau_l$ & $\tau_h$ & $\rho_{\text{strict}}$ & $\rho_{\text{lenient}}$ \\
\midrule
\multirow{4}{*}{ProcessBench} 
  & \multirow{2}{*}{Qwen3-1.7B} & \ourmethoduni & \multirow{2}{*}{\{16, 18\}} & \multirow{2}{*}{3.0} & -- & \multirow{2}{*}{0.6} & -- & \multirow{2}{*}{0.6} & -- \\
  & & \ourmethodbi &  &  & 0.5 &  & 0.7 &  & 0.4 \\
  \cmidrule{2-10}
  & \multirow{2}{*}{Qwen3-8B} & \ourmethoduni & \multirow{2}{*}{\{22, 23\}} & \multirow{2}{*}{\{1.0, 2.0\}} & -- & \multirow{2}{*}{0.5} & -- & \multirow{2}{*}{0.0} & -- \\
  & & \ourmethodbi &  &  & 1.5 &  & 0.7 &  & 0.0 \\
\midrule
\multirow{4}{*}{Hard2Verify} 
  & \multirow{2}{*}{Qwen3-8B-thinking} & \ourmethoduni & \multirow{2}{*}{\{22, 23\}} & \multirow{2}{*}{1.5} & -- & \multirow{2}{*}{0.6} & -- & \multirow{2}{*}{0.0} & -- \\
  & & \ourmethodbi &  &  & 1.5 &  & 0.7 &  & 0.1 \\
  \cmidrule{2-10}
  & \multirow{2}{*}{FARE-20B} & \ourmethoduni  &   \multirow{2}{*}{\{17, 21\}} & \multirow{2}{*}{1.0} & -- & \multirow{2}{*}{0.6} & -- & \multirow{2}{*}{0.4}  & -    \\
  &                           & \ourmethodbi    &   &   &   1.0 &  & 0.7  &  &    0.4                 \\
\bottomrule
\end{tabular}%
}
\caption{Steering hyperparameters for each model and benchmark setting. For Qwen3-8B on ProcessBench, $\alpha_{\text{strict}}$ is tuned per subset: 2.0 on the easier GSM8K and MATH, and 1.0 on the harder OlympiadBench and Omni-MATH.}
\end{table}


\section{Additional Results}\label{app:results}

\Cref{tab:compare-veri-cot-complete} provides the complete set of case studies summarized in \Cref{tab:compare-veri-cot} of the main paper, including two additional examples (one under-critical and one over-critical). All examples use Qwen3-8B on ProcessBench with the basic prompt.

In the second under-critical case (row 2), the baseline verifier blindly repeats the solution's algebraic expansion without independently checking the intermediate steps, concluding that $a^4 - 152a^2 + 196 = 0$ is correct. After steering with $\strictvwol$, the verifier performs its own expansion, arrives at $a^4 - 197a^2 + 196 = 0$, and correctly identifies the discrepancy.
In the second over-critical case (row 4), the baseline verifier acknowledges that the final result is correct but flags the use of an approximate decimal value as an error in ``reasoning and precision.'' After steering with $\lenientvwol$, the verifier recognizes that the approximation is consistent with the exact value $\sqrt{8}$ and correctly accepts the step.

\begin{table}[!ht]
\centering
\scriptsize
\setlength{\tabcolsep}{2pt}
\resizebox{\linewidth}{!}{
\begin{tabular}{p{0.05\linewidth} p{0.22\linewidth} p{0.35\linewidth} p{0.35\linewidth}}
\toprule
& \multicolumn{1}{c}{\textbf{Solution Step}} &
\multicolumn{1}{c}{\textbf{Baseline Incorrect Verification}} &
\multicolumn{1}{c}{\textbf{Steered} ({\setlength{\fboxsep}{1pt}\colorbox{verylightblue}{+$\strictvwol$}}/{\setlength{\fboxsep}{1pt}\colorbox{verylightred}{+$\lenientvwol$}})} \\

\midrule
\multirow{2}{*}[-2.5em]{\rotatebox{90}{\textbf{Under-critical}}}
& \multicolumn{1}{p{0.22\linewidth}|}{\textit{After using 25 bandages, the remaining bandages are \( \textcolor{red}{(x - 16 + 100) - 25} = \textcolor{red}{x + 69 - 25 = x + 44} \).}}
& \multicolumn{1}{p{0.35\linewidth}|}{\textbf{Analysis}: This is \textcolor{red}{correct}. They added 100 bandages, used 25, so the remaining is $ \textcolor{red}{x - 16 + 100 - 25} = \textcolor{red}{x + 69 - 25 = x + 44} $. \textbf{No error}.}
& \cellcolor{verylightblue} \textbf{Analysis}: This paragraph \textcolor{red}{contains an \textbf{error}}. The expression $ \textcolor{red}{(x - 16 + 100) - 25} $ simplifies to \textcolor{red}{$ x + 84 - 25 = x + 59 $, not $ x + 44 $}. The error is in the arithmetic: $ -16 + 100 = 84 $, not 69. So the correct remaining bandages \textcolor{red}{should be $ x + 59 $, not $ x + 44 $.} \\
\cmidrule{2-4}

& \multicolumn{1}{p{0.22\linewidth}|}{\textit{...we get \((49 - a^2)(4 - a^2) = 144a^2\). Expanding and simplifying, we have \(196 - 8a^2 + a^4 = 144a^2\), which simplifies to \textcolor{red}{\(a^4 - 152a^2 + 196 = 0\).}}}
& \multicolumn{1}{p{0.35\linewidth}|}{\textcolor{red}{\textbf{Correct.}} Substituting $b^2 = 49 - a^2$ into the equation $b^2(4 - a^2) = 144a^2$ gives $(49 - a^2)(4 - a^2) = 144a^2$. Expanding and \textcolor{red}{simplifying leads to $a^4 - 152a^2 + 196 = 0$.}}
& \cellcolor{verylightblue} Analysis: Let's check the substitution and expansion:- Substitute (...). \textcolor{red}{So the equation becomes:
$196 - 53a^2 + a^4 = 144a^2$ \textnewline \, $a^4 - 197a^2 + 196 = 0$ - But the solution says $a^4 - 152a^2 + 196 = 0$}, which is \textcolor{red}{incorrect}. \\
\midrule

\multirow{2}{*}[-4.5em]{\rotatebox{90}{\textbf{Over-critical}}}
& \multicolumn{1}{p{0.22\linewidth}|}{\textit{First, calculate the area of the larger rectangle. The larger rectangle has a length of 4 yards and a width of 5 yards. \([ \text{Area of the larger rectangle} = 4 \times 5 = 20 \text{ square yards} ]\)}}
& \multicolumn{1}{p{0.35\linewidth}|}{\textbf{Analysis}: ...However, the dimensions are not clearly defined in the diagram, and the interpretation of which dimension corresponds to which side is ambiguous. - \textbf{Conclusion}: \textcolor{red}{The error is in the \textbf{interpretation of the dimensions}}, ...}
& \cellcolor{verylightred}  This is also correct. The diagram labels ``4 yds'' at the top and ``5 yds'' on the left side, so the larger rectangle is 4 yards wide and 5 yards tall. The area is indeed $4 \times 5 = 20$ square yards. \\
\cmidrule{2-4}

& \multicolumn{1}{p{0.22\linewidth}|}{\textit{We are asked to find \(x^2\). Thus, we calculate: \([ x^2 = (2.8284)^2 = 8 ]\)}}
& \multicolumn{1}{p{0.35\linewidth}|}{Analysis: This is incorrect. ... While ..., the reasoning is slightly misleading because it uses an approximate decimal value instead of an exact expression.
However, the final result is still correct. \textcolor{red}{But the earliest error is in the reasoning and precision of the calculation}.}
& \cellcolor{verylightred} This is also correct. $(2.8284)^2$ is approximately $8$, and since the coordinates of $E$ were given with high precision, this suggests that the exact value is $\sqrt{8}$, and thus $x^2 = 8$. \\
\bottomrule
\end{tabular}}
\caption{\textbf{Representative case studies of verification failures and their correction by hidden-state steering.} Qwen3-8B on ProcessBench. The top two rows show under-critical cases, where the baseline verifier falsely accepts an erroneous solution step, and the bottom two rows show over-critical cases, where it falsely rejects a correct step. The ``Steered'' column shows that applying $\strictvwol$ corrects false acceptances into true rejections, while applying $\lenientvwol$ corrects false rejections into true acceptances.
}
\label{tab:compare-veri-cot-complete}
\end{table}

\end{document}